%% file: main.tex
\documentclass[letterpaper, 10 pt, conference]{ieeeconf}  %
\usepackage{graphicx}
\usepackage{color}
\usepackage{subfigure}
\usepackage{url}
\usepackage{amsmath}
\usepackage{comment}
\usepackage{cite}
\usepackage{float}
\usepackage{marvosym}

\title{\LARGE \bf
A Bio-Inspired Tensegrity Manipulator with Multi-DOF, Structurally Compliant Joints
}

\author{Steven Lessard$^{1,2}$, Dennis Castro$^{1}$, William Asper$^{1}$, Shaurya Deep Chopra$^{1,3}$, \\ Leya Breanna Baltaxe-Admony$^{1}$, Mircea Teodorescu$^{1,3}$, Vytas SunSpiral$^{2,4}$, and Adrian Agogino$^{1,2}$%
\thanks{$^*$This material is based upon work supported by the National Aeronautics and Space Administration under Prime Contract Number NAS2-03144 awarded to the University of California, Santa Cruz, University Affiliated Research Center.}
\thanks{$^{1}$University of California, Santa Cruz, Santa Cruz, CA 95064}
\thanks{$^{2}$Authors with the NASA Ames Dynamic Tensegrity Robotics Lab, Moffett Field, CA 94035}
\thanks{$^{3}$NASA-ARC Advanced Studies Laboratories, Moffett Field, CA 94035}
\thanks{$^{4}$Stinger Ghaffarian Technologies, Greenbelt, MD 20770, USA}
}

\begin{document}

\maketitle
\thispagestyle{empty}
\pagestyle{empty}

\input{abstract}


\input{background}

\input{simulation}

\input{sysdesign}

\input{results}

\input{conclusion} 

\input{acknowledgment}


\bibliographystyle{IEEEtran}
\bibliography{tensegrity}

\end{document}

%% file: abstract.tex
\begin{abstract}
Most traditional robotic mechanisms feature inelastic joints that are unable to robustly handle large deformations and off-axis moments.
As a result, the applied loads are transferred rigidly throughout the entire structure.
The disadvantage of this approach is that the exerted leverage is magnified at each subsequent joint possibly damaging the mechanism.
In this paper, we present two lightweight, elastic, bio-inspired tensegrity robotic arms adapted from prior static models which mitigate this danger while improving their mechanism's functionality.
Our solutions feature modular tensegrity structures that function similarly to the human elbow and the human shoulder when connected. 
Like their biological counterparts, the proposed robotic joints are flexible and comply with unanticipated forces.
Both proposed structures have multiple passive degrees of freedom and four active degrees of freedom (two from the shoulder and two from the elbow).
The structural advantages demonstrated by the joints in these manipulators illustrate a solution to the fundamental issue of elegantly handling off-axis compliance.
Additionally, this initial experiment illustrates that moving tensegrity arms must be designed with large reachable and dexterous workspaces in mind, a change from prior tensegrity arms which were only static.
These initial experiments should be viewed as an exploration into the design space of active tensegrity structures, particularly those inspired by biological joints and limbs.
\end{abstract}

%% file: background.tex
\section{INTRODUCTION}
\begin{figure}[!ht]
	\label{main} 
    \centering 
    \subfigure[Arm]{ \includegraphics[width=.75\linewidth]{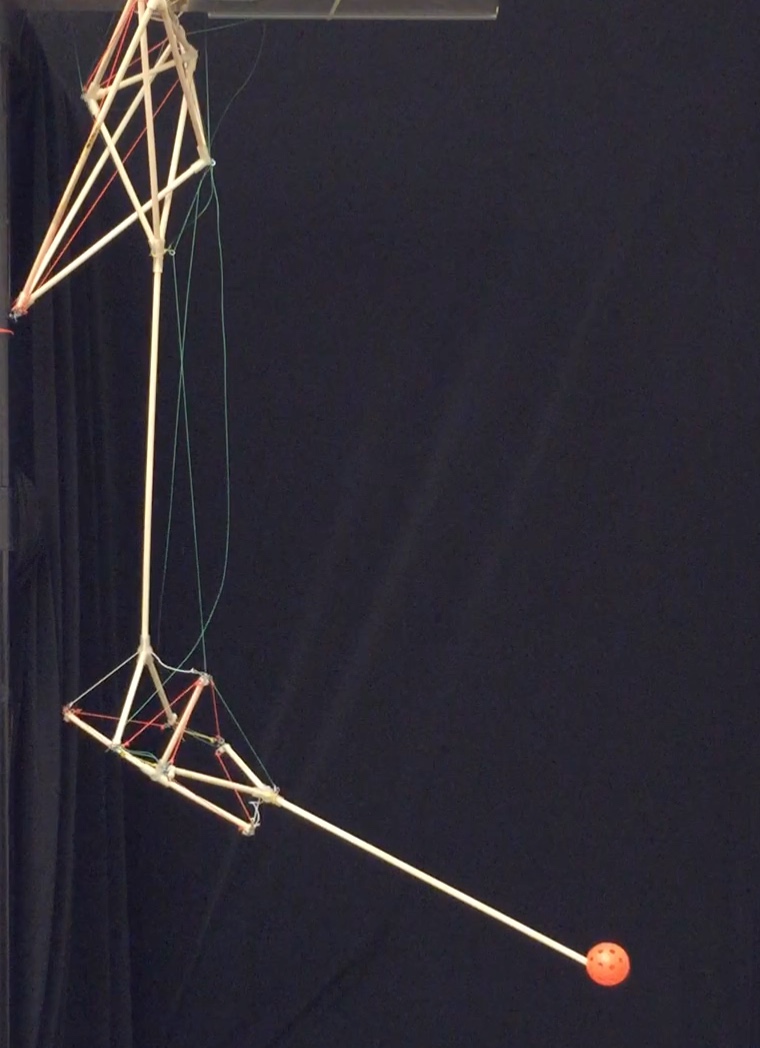}} 
	\subfigure[``Elbow'']{\includegraphics[height=.45\linewidth]{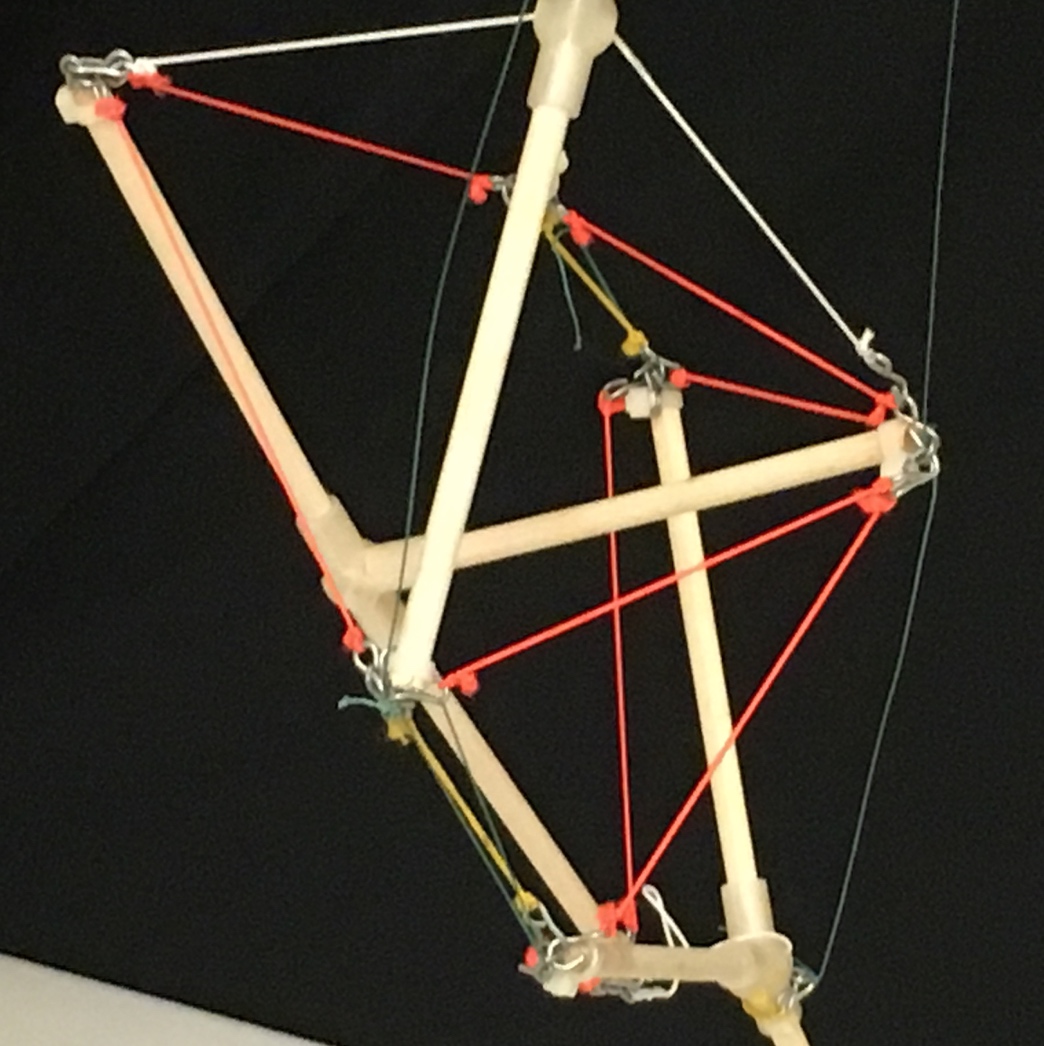}} 
	\subfigure[``Shoulder'']{\includegraphics[height=.45\linewidth]{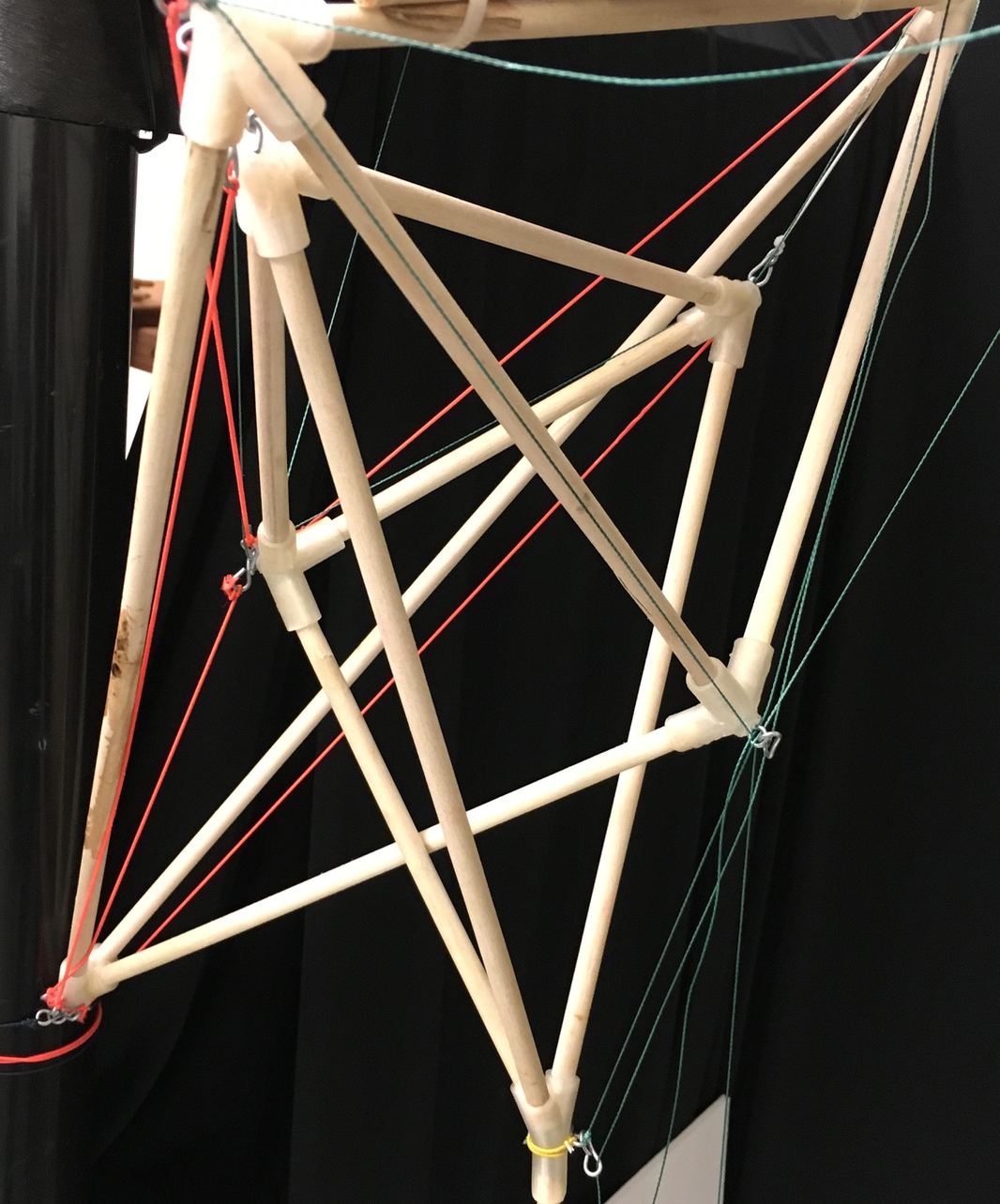}}    
\caption{A tensegrity manipulator featuring two multi-DOF joints. This model uses the tetrahedrons design for its shoulder joint and the elbow joint first described by Lessard et al. \cite{lessard2016compliantjoint}} 
\end{figure}

The structure of the human body is largely defined by the span of its bones, muscles and connective tissues. 
Bones act as compression elements that rigidly define the general shape and are capable of directly supporting loads.
Muscles act as tension elements, which actuate the bones to transmit motion and force, while the connective tissue envelops the bones and muscles acting as a compliant and restrictive medium between the two. 
The result of this heterogeneous mixture is a system that can articulate in many degrees of freedom while protecting itself from impacts.

The ability to simultaneously articulate in many degrees of freedom while protecting itself from impacts has been a central topic of robotics research.
As a result, many anthropomimetic robots have been created to take advantage of the mimicking of human body.
The cable driven Anthrob robot features a bone themed compressive architecture that supports a cable-driven matrix of ``muscles"\cite{jantsch2013anthrob}.

Other cable-driven robots, like Kenshiro, features the integration of muscle and bone facsimiles \cite{nakanishi2012design} of an entire body.
Boasting hundreds of actuators, Kenshiro illustrates the workspace and precision in articulation a humanoid robot can achieve.
As the need for increasingly complex controllers for these robots has grown, the research into the mathematical representation of these complex and often stochastic systems has also become increasingly relevant.
D. Lau et al \cite{lau2013generalized} showed how the dynamics and kinematics of a system can be expressed through a tensile adjacency matrix.
In these matrices, indirect action and actuation can be derived from initial conditions, allowing for more precise and accurate control. 
Beyond cable based actuation, the ability of a robot to protect itself from impacts can also be aided by the use of compliant actuators like dielectric elastomers \cite{pelrine2002dielectric} or pneumatic based McKibben air muscles\cite{tondu2012modelling}. 
 
In addition to biomimetic material and structural approaches, there have been successful attempts to model the kinematics and dynamics of human motion through robots as well.
Khatib et al. used direct marker tracking to illustrate how human movements, especially highly dynamic athletic ones, can be characterized and modeled \cite{khatib2009robotics}.
To address the concern for realistic movement during human manipulation tasks, Zanchettin et al. described an approach to observe and recreate human-like movement using kinematic redundancy \cite{zanchettin2011kinematic}.
The culmination of works such as these has established a foundation upon which future biomimetic humanoid robots can be compared.
 


Traditional designs that do not consider the mechanical compliance at the structural level fail to address external off-axis forces.
Our solution to this problem, tensegrity (``\textit{tensile-integrity}") structures, combine soft and rigid systems with the benefits of both traditional designs and structural compliance. 

Tensegrity structures are composed of compression elements suspended within a matrix of tension elements.
Passive tensegrity structures, such as those constructed by Tom Flemons, Stephen Levin, and Graham Scarr, have previously implemented human limbs \cite{cretu2009tensegrity, levin2002tensegrity, scarr2012consideration}.
These structures, as well as investigations led by Turvey and Fonseca \cite{turvey2014medium}, have cited the need for the construction of active tensegrity models to study the biomechanics of human limbs, especially the arm.

Active tensegrity structures have been used to build modular substructures to achieve biomechanical motion.
Mirletz et al. used tetrahedral links to create a locomoting spine \cite{mirletz2014design}.
Lessard et. al designed a modular 2 degree-of-freedom tensegrity joint to emulate the human elbow \cite{lessard2016compliantjoint}.
These active structures, like their passive predecessors, were inspired by and used to abstractly model the anatomy and function of corresponding biological structures

In this paper, we discuss how these motivating factors influence our design of a tensegrity manipulator (Figure 1).
First, we illustrate how the simulation of tensegrity structures predicts how physical prototypes will behave.
Next, we discuss the biological considerations made when selecting prototypes for construction.
We then demonstrate our physical prototypes and report on the observed capabilities of these robots.
After discussing the implications of our active tensegrity structure, we conclude with a summary of our findings, some practical applications of the technology as well as the direction of future work.

%% file: simulation.tex
\section{SIMULATION}
\begin{figure}[ht]
	\label{anatomyvsrobot} 
    \centering
    \includegraphics[height=.5\linewidth]{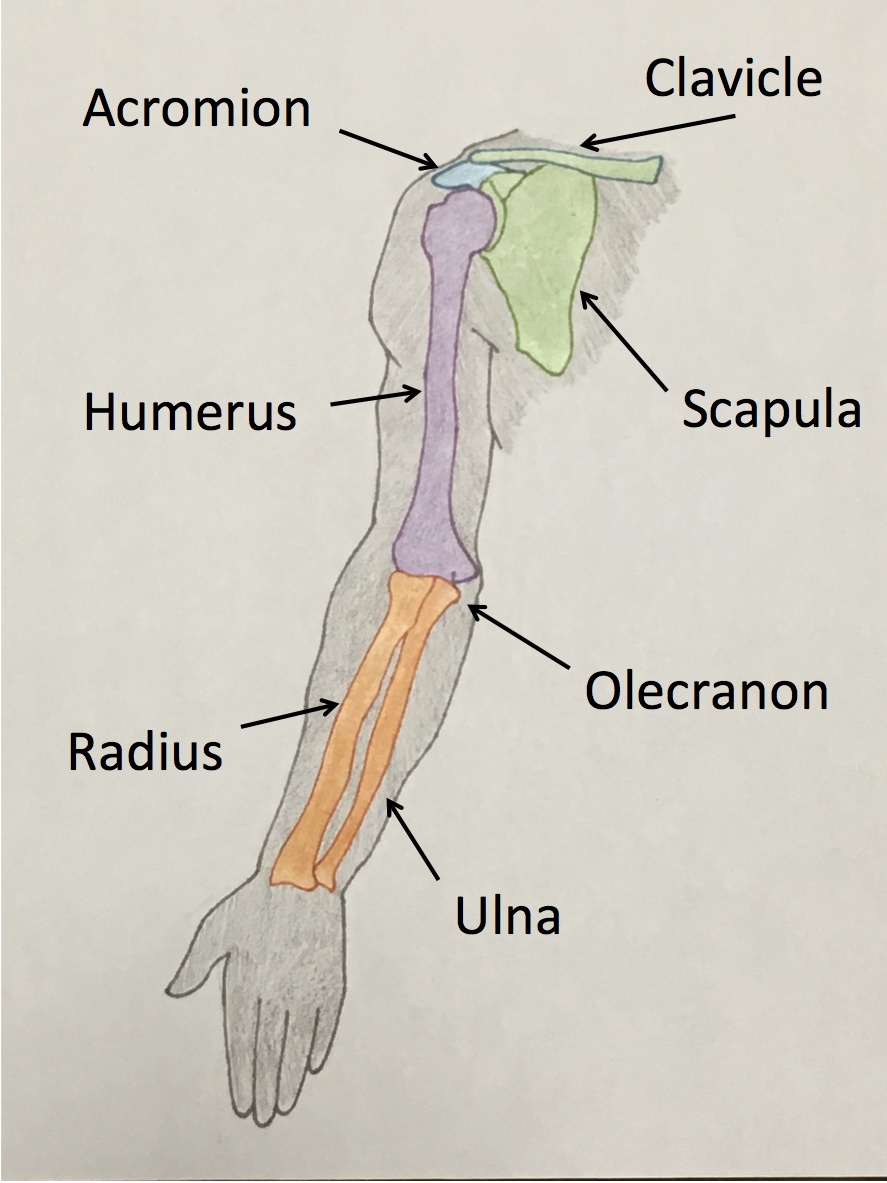}
    \includegraphics[height=.5\linewidth]{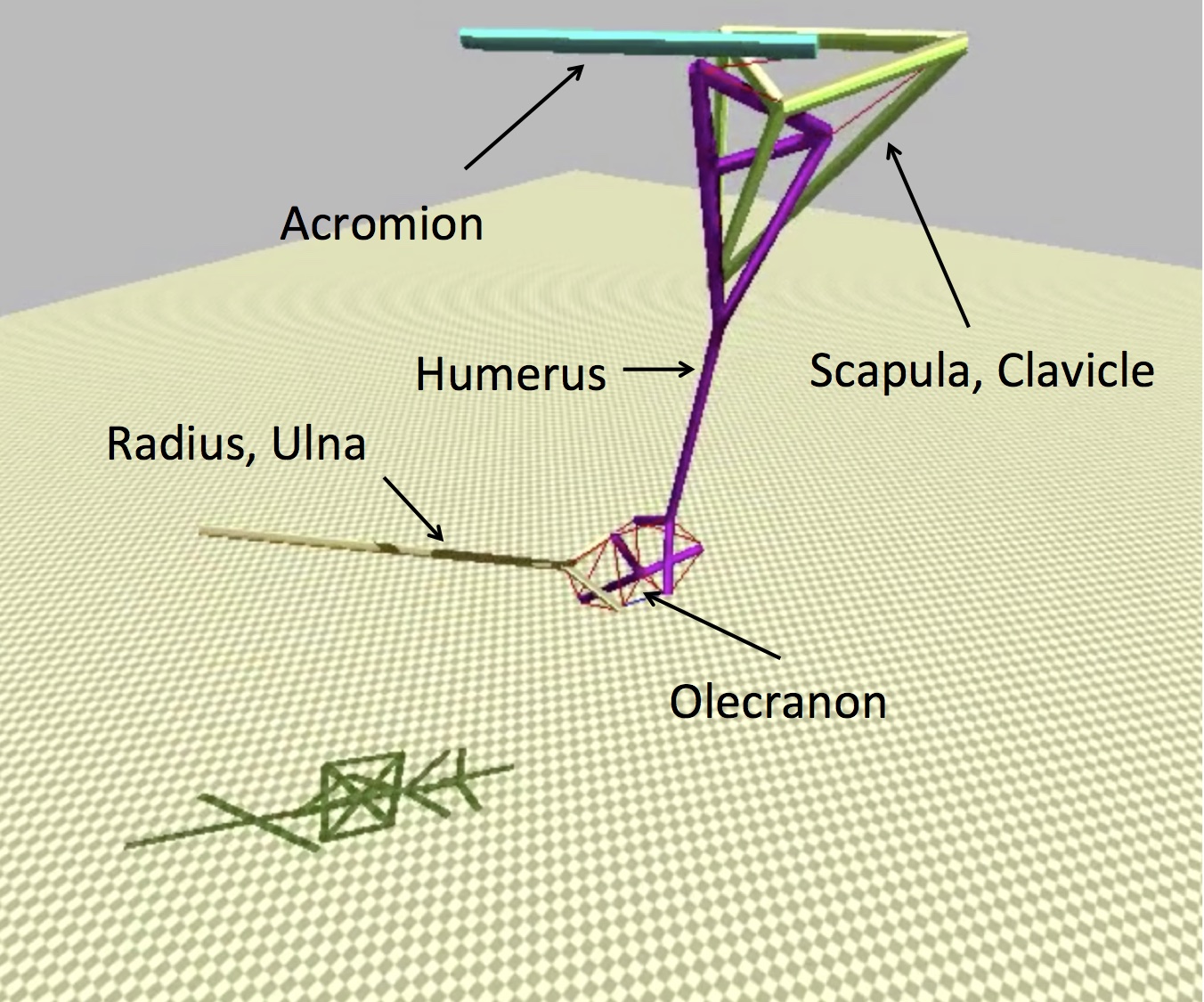}
    \caption{A mapping between the major compression elements of the human arm (left) and those of the simulated tensegrity manipulator models.} 
\end{figure}

To determine which physical prototypes were feasible robotic arms, we initially simulated an array of various tensegrity robotic arms with NASA's Tensegrity Robotics Toolkit (NTRT).
NTRT is an open-source simulator that allows users to design and control tensegrity structures and robots which has been built on top of the Bullet Physics Engine (version 2.82)\footnote{Additional information about NTRT can be found at \\ \url{http://irg.arc.nasa.gov/tensegrity/NTRT}}. 
These simulations illustrated which models were stable when controlled by a periodic signal.
 

Earlier investigations found potential models that can be adopted for the design of a tensegrity shoulder.
One design, initially based upon the DuCTT robot \cite{friesen2014ductt}, uses two interlocked tetrahedrons for climbing duct systems.
We adapted this design to handle both translational and rotational degrees of freedom similar to that of the human arm.
An NTRT simulation of this joint \cite{baltaxe2016shoulder} provided a clearer use for this configuration on a bio-inspired manipulator.
We have built and simulated a design upon the amalgamations of these existing models within NTRT (Figure 2). 

The original simulations on tensegrity shoulder joints by Baltaxe-Admony, Robbins et al. were developed and used to test and compare range of motion and degrees of freedom in the manipulator \cite{baltaxe2016shoulder}. 
These findings illustrated how four total active degrees of freedom could be generated in the hardware prototype.
\begin{align}
F = -kX - bV \label{hookeslawdamping}
\end{align}
Within the simulator, cables are modeled as two connected points whose medium length changes according to Hooke's Law for linear springs with a linear damping term as well (Equation \ref{hookeslawdamping}).
Cable control is dictated by functions within a controller class, meaning that the exact length of the cable can be set at each time step according to a control policy.
Real-world limitations, such as the max acceleration of the motor used and the target velocity of cable lengthening are added to the simulation as well at the structural level.
In addition, maximum and minimum lengths can be applied to each individual cable to prevent unnatural deformations.
These features assert that the robot in simulation is never given extraordinary means to accomplish its goal.
The use of NTRT has already been shown in previous papers to have produced accurate statics and dynamics for SUPERBall, a tensegrity rover designed for extraterrestrial missions \cite{caluwaerts2014design, mirletzcpgs}.

%% file: sysdesign.tex
\section{MODEL SELCTION}
In an effort to recreate the structural compliance and flexibility observed in human arms, we base our designs for the tensegrity arm upon an abstracted anatomical model of the underlying musculoskeletal system and connective tissue (i.e. fascia, tendons, and ligaments) in biological arms.
Compression elements are oriented and situated to mimic bones or groups of bones and tension elements were arranged to connect the compression elements like connective tissue.
The true anatomy of the arm is not replicated in this tensegrity model because of its sheer complexity.

Designs that emulate these principles have been simulated previously by Baltaxe-Admony, Robbins, et al. \cite{baltaxe2016shoulder}.
From their results, we chose to implement the complex saddle and tetrahedrons models.
Our tensegrity models focus on the minimum number of components required to obtain the degrees of freedom and workspace for a two-joint manipulator.
The two models chosen illustrated this desired behavior while demonstrating the largest reachable workspaces.


\section{SYSTEM DESIGN}
The construction of the two selected models (the saddle arm and the tetrahedrons arm) can be divided into two parts: the compression elements and the tension elements.

\subsection{Compression Elements}
The compression elements of the actuated tensegrity arm are primarily constructed out of wood (Figure 1).
Wood is lightweight but still rigid enough to prevent buckling and plastic deformation due to the pull from connecting tension elements.
Wooden rods were connected to one another with 3D-printed Polylactic Acid (PLA) end caps.
\par The nexuses observed in human arms inspired the functionally similar tensegrity joints in our model.
These arm models use the same design as the elbow joint first described by Lessard et al. \cite{lessard2016compliantjoint} which functions similarly to a Cardan joint.
In this elbow design, the number of compression and tension elements are minimized while maintaining two active degrees of freedom (and many more redundant passive degrees of freedom).
As a result, the radius and ulna are merged into one compression element and the olecranon (the hook of the ulna) is a separate element. 
The olecranon serves as a hub to route actuation from the top of the arm through the humerus and to the forearm.

The design of the two shoulder joints follow a similar philosophy.
The humerus compression element connects to an upper shoulder compression element via tension elements.
The minimalist format of the tetrahedrons shoulder emulate the human scapula, clavicle and proximal end of the humerus (Figure 1).
The acromion is a separate compression element that supports the rest of the tetradehons arm.
In the saddle shoulder, the proximal end of the humerus mimics the distal end using a ``y-connector".
An additional perpendicular ``y-connector'' intercepts said proximal end of the humerus to create a point about which the humerus can swivel (Figure 4).
When combined with the tensegrity elbow, both the tetrahedrons arm and the saddle arm create four degrees of motion (Table II). 

\subsection{Tension Elements}
Tension elements in a tensegrity structure hold the compressive skeleton of that structure together at equilibrium.
In this model, the tension elements are one of two varieties: passive or active.

\begin{figure}
	\label{TensionAnatomy}
    \centering
    \includegraphics[width=0.4\linewidth]{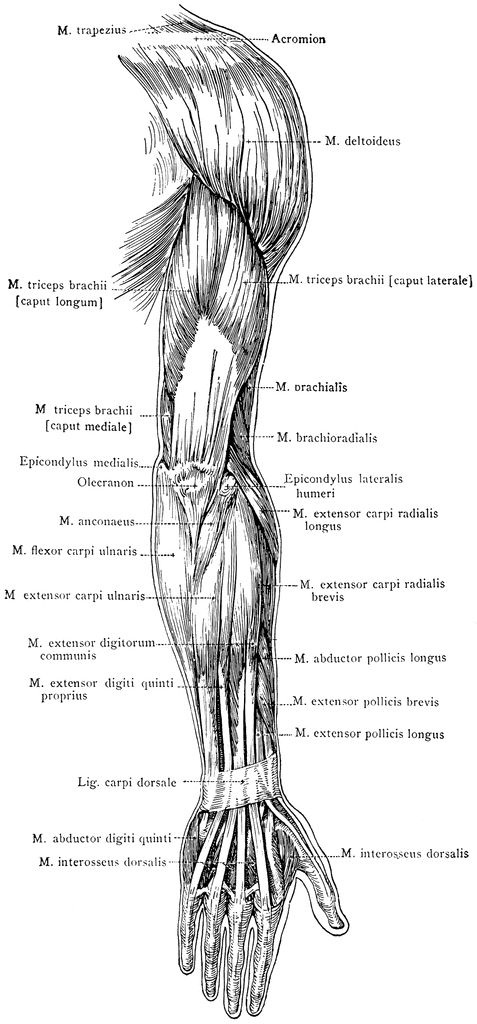}
    \includegraphics[width=0.4\linewidth]{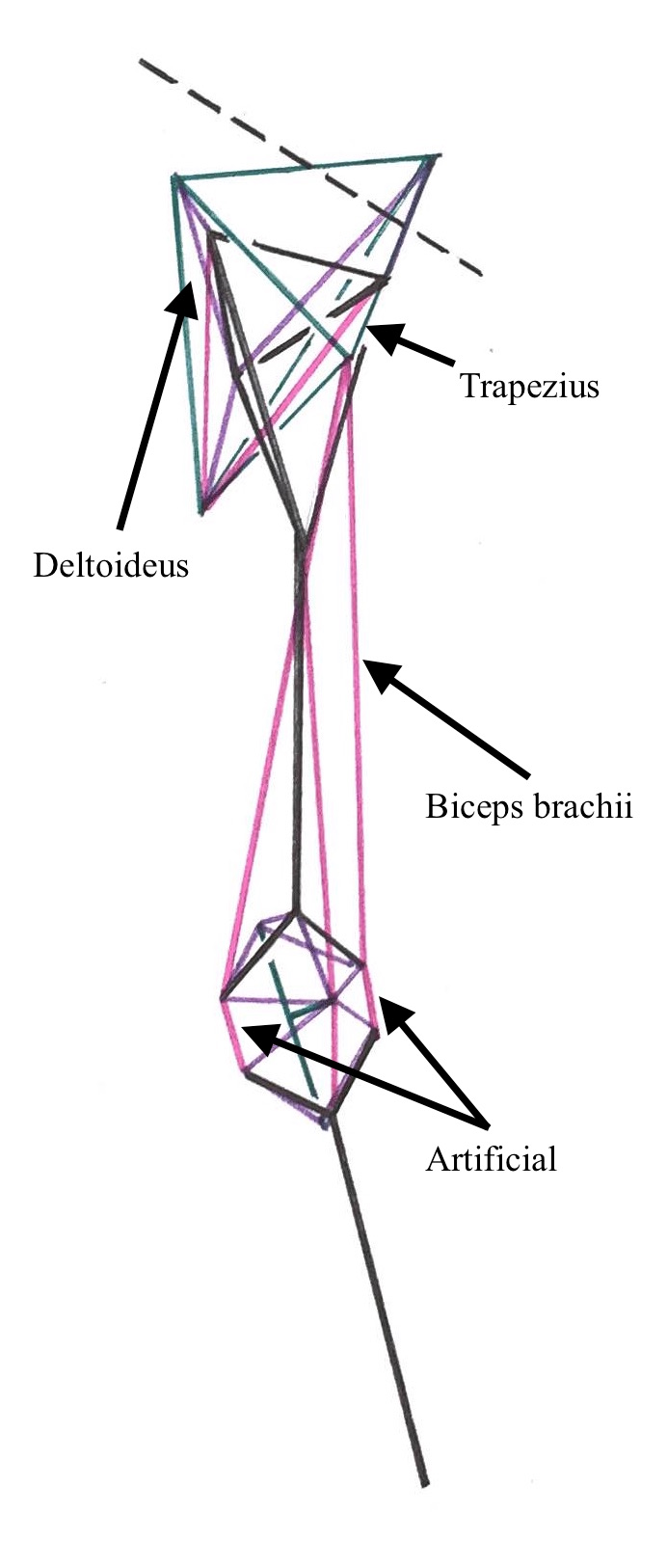}
    \caption{Posterior view of the superficial muscles of the human arm (from Gray's Anatomy $41^{st}$ edition) and a corresponding view of the proposed tetrahedrons tensegrity robotic manipulator. The cables (purple) indicated by the arrows are the actuated cables in the tetrahedrons prototype.}
\end{figure}

\subsubsection{Passive Tension}
Passive tension elements are not directly actuated and provide static support in the structure.
They are especially important for handling unanticipated forces, both internal and external.
Their elasticity allows the overall structure to temporarily absorb impacts and deform while the applied load is distributed throughout the other elements of the structure.
Our model uses parachute cord (paracord) for the passive tension elements because it is light and elastically stretches easily. 
As a result, the passive tension elements of our model are most similar functionally to the connective tissue in the human body.

\subsubsection{Active Tension}
Active tension elements are tension elements which are directly actuated.
Unlike passive tension elements, active tension elements require strong axial strength to bear loads applied along their intended degree of freedom.
Our model uses an aramid-based fishing line for the active tension elements, since it is comparably very strong at lifting large masses.
Functionally, the active tension elements in our model abstractly replicate muscles in the human body.
One important difference between the two is the number of parallel fibers.
In true muscles, there are thousands of fibers pulling in unison to flex; our model however features a single fiber instead.
Theoretically, more fibers in parallel will increase the actuation strength, but at the cost of requiring more power and more motors.

Tension elements in the two implementations provided are strung through small metal hoops attached to the vertices of the wooden elements.
Because the motors for the arm are located above the arm itself, tension elements are anchored at their distal ends.
As the motors are run, the corresponding active tension elements are wound or unwound around an attached spool.

%% file: results.tex
\section{PRELIMINARY DESIGNS AND RESULTS}
In this section, we compare a constructed tensegrity saddle arm with a constructed tensegrity tetrahedrons arm (Figures 1 and 4).
Most notably in this section, we describe the exact parameters that allow a tensegrity manipulator to actuate and move similarly to a human arm.
The resulting theory is important for translating static tensegrity arm designs into actuated tensegrity robotic prototypes.

The first shoulder joint (Figure 1) features nested tetrahedrons and the second design (Figure 4) features a saddle joint.
The two designs contrast in the degrees of freedom they offer: the nested tetrahedrons model offers lift as opposed to the yaw the saddle model creates.

In Tables I and II, we list the lengths and masses of the compression elements in the tetrahedrons and saddle models.
Note that the "Full Arm" length is measured as from the end of the forearm to the top of the shoulder when the arm is completely elongated (i.e. biceps cable extended, triceps cable contracted). 
Individual component lengths are independent of this elongation (they are measurements of just the compression elements). 
This table excludes the mass of the components, which are not supported by the tensegrity structure (e.g. motors, control boards, and power supply).

\begin{table}[H]
\label{armmetrics_1}
\caption{Tetrahedrons Arm Measurements}
\vspace{3 mm}
\begin{center}
\begin{tabular}{ | c | c | c | }
 \hline
Component & Mass (g) & Length (cm) \\ 
 \hline
 Forearm & 10.1 & 58.6 \\  
 Olecranon & 6.0 & 24.0 \\
 Humerus & 36.6 & 76.2 \\
 Shoulder Tetrahedron & 24.8 & 36.1 \\
 \hline
 Full Arm & 77.5 & 149 \\
 \hline
\end{tabular}
\end{center}
\end{table}

\begin{table}[H]
\label{shoulderJoint_1}
\caption{Saddle Arm Measurements}
\vspace{3 mm}
\begin{center}
\begin{tabular}{ | c | c | c | }
 \hline
 Component & Mass (g) & Length (cm) \\
 \hline
 Forearm & 18.5 & 54 \\
 Olecranon & 11.6 & 24 \\
 Humerus & 23.2 & 53\\
 Saddle Joint & 36.1 & 54 \\
 \hline
 Full Arm & 89.4 & 104 \\
 \hline
\end{tabular}
\end{center}
\end{table}

\begin{figure}[ht]
\label{saddleshoulder} 
    \centering
    \includegraphics[width=.9\linewidth]{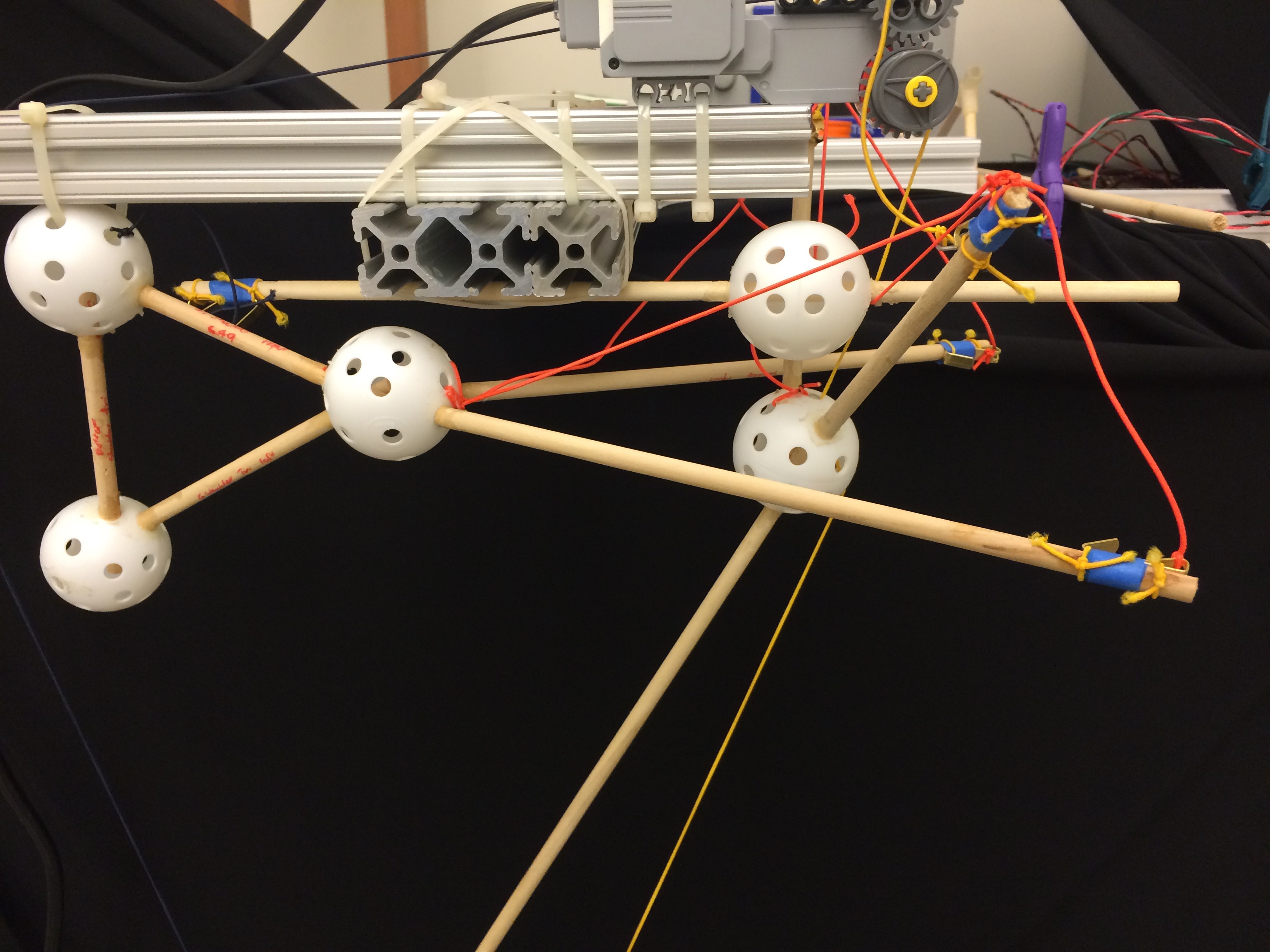}
    \caption{The saddle shoulder joint features two perpendicular ``y-connectors" held together in tension. The combination of these two compression elements demonstrate yaw (shoulder rotation).}
\end{figure}

Each of our hardware prototypes feature four active degrees of freedom and many passive degrees of freedom.
Using a microcontroller with an attached motor shield, we actuated four DC motors to articulate the arm. 
The motors in each model are offloaded to a platform above the shoulder joint from where the remaining tensegrity components hang. 
To illustrate the full workspace of the manipulator, we used a controller that wound and unwound each cable according to a sinusoidal wave.

Each actuated rotation by the manipulators uses two tension elements which are invariably inversely proportional in length.
In this manner, these antagonistic pairs of tension elements emulate the dynamic observed in many muscle groups (e.g. biceps flexion/triceps extension and triceps flexion/biceps extension).

\begin{table}[H]
\label{dof}
\caption{Degrees of Freedom in Tensegrity Manipulators}
\vspace{3 mm}
\begin{center}
\begin{tabular}{ | c | c | c | }
 \hline
 Joint & Motion & Abstracted Biological Analog \\
  \hline
  \multicolumn{3}{|c|}{Tetrahedrons Shoulder Joint} \\
  \hline
  Shoulder & Pitch & Forward shoulder raise \\
  Shoulder & Lift & Shrug \\
  Elbow & Pitch & Biceps curl \\
  Elbow & Yaw & Artificial \\
  \hline
  \multicolumn{3}{|c|}{Saddle Shoulder Joint} \\
  \hline
   	Shoulder & Pitch & Forward shoulder raise \\
 	Shoulder & Yaw & Internal/External shoulder rotation \\
 	Elbow & Pitch & Biceps curl \\
 	Elbow & Yaw & Artificial \\
 \hline
\end{tabular}
\end{center}
\end{table}


\begin{figure}[ht]
	\label{compliance_sim} 
    \centering
    \subfigure[Simulated trajectory of the end effector (orange ball in (b)) hitting a rigid obstacle during pitch motion] {\includegraphics[width=.85\linewidth]{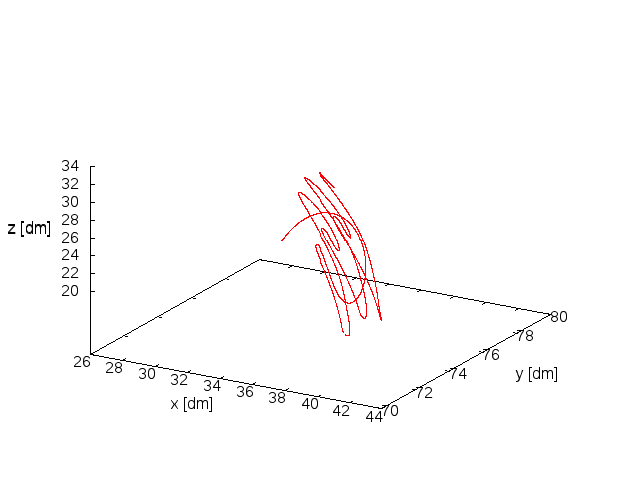}}
    \subfigure[Physical representation of the simulation shown in (a)] {\includegraphics[width=.85\linewidth]{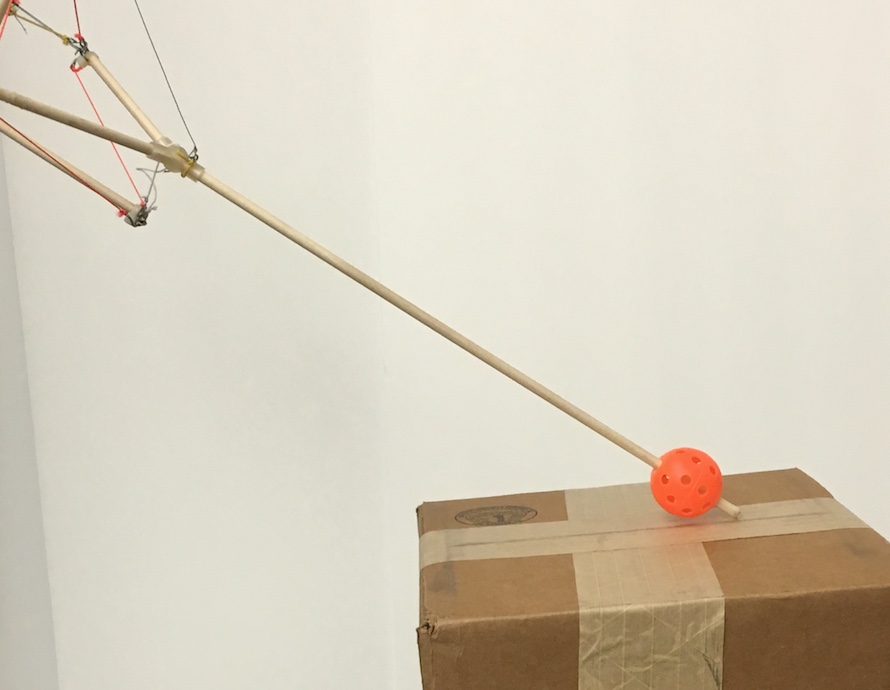}}
    \caption{Trajectory of the end-effector as it collides with a rigid object during pitch motion} 
\end{figure}

\begin{figure}[ht]
	\label{stress_limits} 
    \centering
    \textbf{}\par\medskip
    \begin{table}[H]
        \label{rom}
		\caption{Flexibility Strain Limits}
		\vspace{3 mm}
		\begin{center}
		\begin{tabular}{ | c | c | }
 		\hline
		Movement & Range of Motion \\ 
 		\hline
 		Elbow Pitch & 215 $^{\circ}$ \\
        Elbow Inward Compression & 2.6 cm \\   
        Elbow Yaw & 40 $^{\circ}$ \\
 		\hline
		\end{tabular}
		\end{center}
	\end{table}
\end{figure}

To handle unexpected external forces, our manipulator features many redundant passive tension elements.

These range of motion values are derived from the minimum and maximum positions of the joints. 
The elbow can contract to a minimum of $35^{\circ}$ to a maximum extension of $250^{\circ}$. 
The elbow can flex along the major axis between a minimum of $23.5 cm$ and a maximum of $26.1 cm$.

These cables absorb and distribute forces throughout the structure, transferring the unanticipated mechanical energy into kinetic and potential energy (swaying and spring compression).
The effects of this behavior can be seen in both the trajectory of the simulated arm (Figure 5a) and the measured flexibility in the hardware prototype (Figure 5b).
We confirmed our simulation findings in the actuated prototype by colliding the end effector of our manipulators with a rigid object (Figure 5b).

\begin{figure}[t]
	\label{motion_sim} 
    \centering
    \subfigure [Tetrahedrons: A: Shoulder Pitch, B: Elbow Pitch, C: Right Yaw, D: Left Yaw, E: Shoulder Lift] {\includegraphics[width=0.7\linewidth]{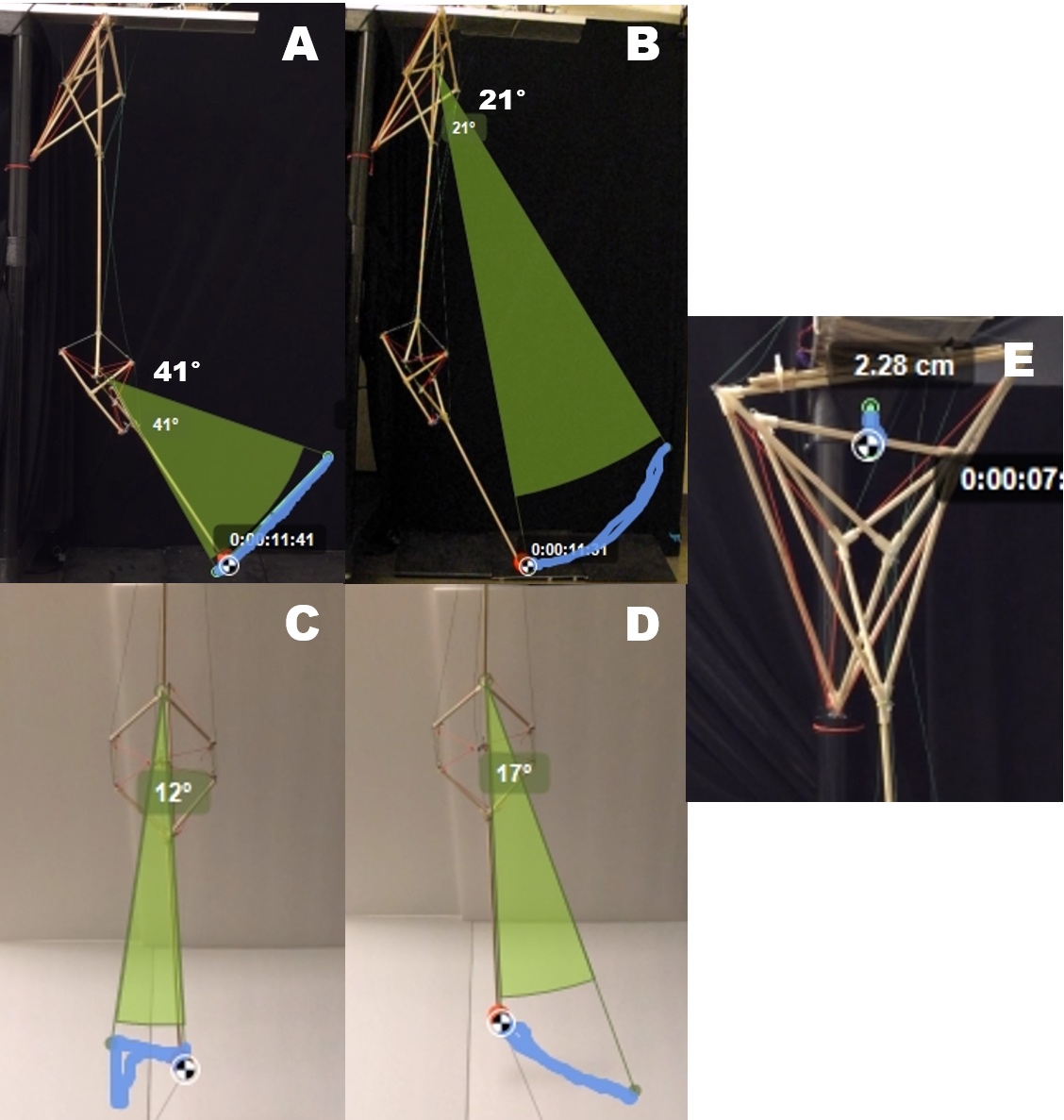}}
    \subfigure[Saddle Joint: A: Elbow Pitch, B: Arm Yaw]{\includegraphics[width=0.8\linewidth]{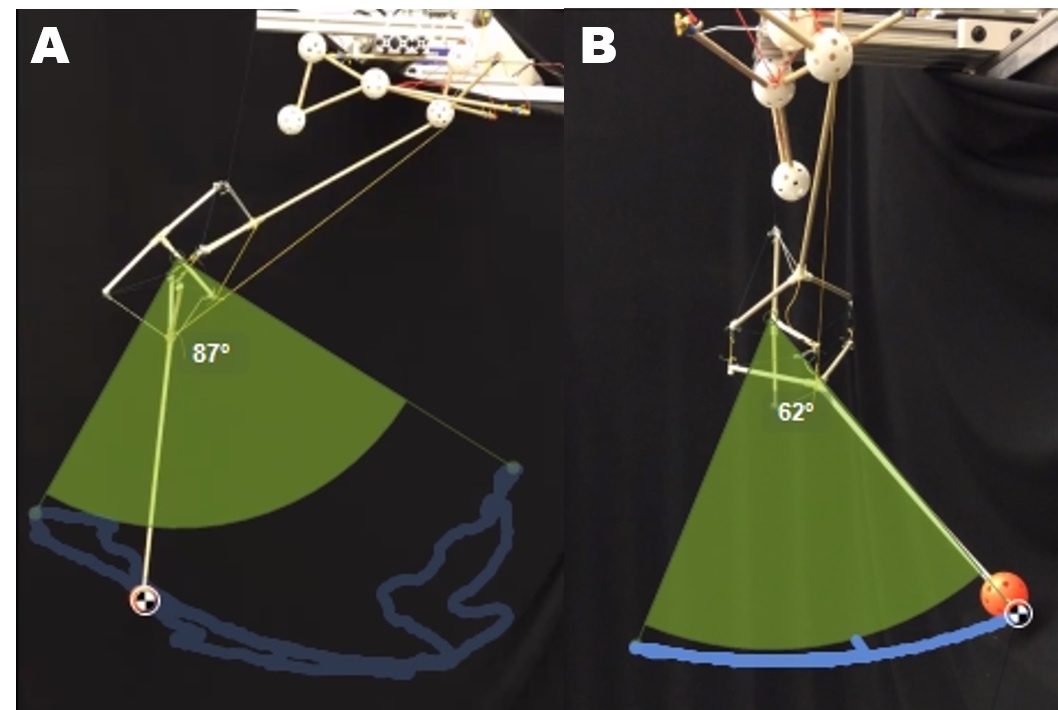}}
    \caption{The tracked motion of the end-effector or part of interest on the tetrahedrons model. The green shaded area represents the angles achieved and the blue lines represent the tracked motion.} 
\end{figure}

\begin{figure}[H]
	\label{compliance_actuated} 
    \centering
    \textbf{}\par\medskip
    \begin{table}[H]
        \label{model_data}
		\caption{Tensegrity Joint Degree of Freedom Flexibility}
		\vspace{3 mm}
		\begin{center}
		\begin{tabular}{ | c | c | c | }
 		\hline
		Movement & Range of Motion & \begin{tabular}{@{}c@{}}Standard \\ Deviation\end{tabular} \\
 		\hline
 		Elbow Pitch & 36.33 $^{\circ}$ & 5.03 $^{\circ}$ \\
        Elbow Yaw (Left)& 14.75 $^{\circ}$ & 2.63 $^{\circ}$ \\
        Elbow Yaw (Right) & 12.00 $^{\circ}$ & 1.414 $^{\circ}$ \\
        Shoulder Pitch & 21.00 $^{\circ}$ & 0 $^{\circ}$ \\
        Shoulder Lift & 2.10 cm & 0.368 cm \\
 		\hline
		\end{tabular}
		\end{center}
	\end{table}
\end{figure}

We tracked the workspace for the tensegrity manipulators seen in Figures 1 and 4, respectively (Figures 6a and 6b). 
Because the two models differ in geometry, these exact workspaces differ between models. 
Each motion of the nested tetrahedron manipulator was then repeated three times to get data points to estimate the range of motion and standard deviation (Table \ref{compliance_actuated}). 
From this we see that the shoulder pitch has a low standard deviation which translates to the transfer of forces only from the base of the model (scapula) to the base of the humerus head. 
Since this is a simple attachment with no external forces acting on the joint the motion is very repeatable. 
In contrast, we see relatively high amounts of deviation between runs for the pitch of the elbow. 
This may be due to the reactive force (extension at the shoulder) needed to ensure pitch at the elbow (pitch at the radius/ulna) transferring throughout the manipulator and causing inconsistent motion.

%% file: conclusion.tex
\section{DISCUSSION, CONTRIBUTION, AND CONCLUSION}
The created tensegrity robots demonstrate that modular structurally compliant joints can form a flexible manipulator.
These robotic joints function according to an abstracted anatomical model of the human arm.
Our tetrahedrons model operates with four active degrees of freedom to create a relatively large workspace.
Our other model, the saddle shoulder, offers an alternative shoulder joint that features yaw motion substituted for the shrug motion from the tetrahedrons model.
This substitution illustrates a trade-off between kinematic redundancy (shoulder model) and more biomimetic behavior (tetrahedrons).
Together, these designs provide important advantages that can improve how future manipulators operate and handle off-axis external forces.

Additionally, these new designs, coupled with empirical implementations, create new insight on the design of translating static tensegrity structures into actuated, moving tensegrity robots.
Previously, static tensegrity structures were used to describe the human joints and limbs (which are inherently active).
Our proposed models illustrate a novel manner for creating active, bio-inspired tensegrity robots.
Thus, these initial experiments explore the design space and should be viewed as early results that will help lead the field towards an active tensegrity design theory.

Manipulators such as these also have a potentially widespread use within other fields of robotics.
In human-oriented systems such as exoskeletons and other wearable robots, our proposed active structures offer a novel method for handling the unique and complex motions of biological systems.
Where traditional robots are over constrained by their rigid structures, tensegrity structures allow for variance in movement, an invaluable attribute in the field of physical therapy. 

Future goals of this work center around extending the principles observed in these tensegrity manipulators and the new active tensegrity design theory for biomimetic limbs and joints.
Exploring alternative materials and actuators for these arms could uncover exact methods that further improve the compliance and controllability.
The creation of a more biologically accurate model and design theory can better explain the exact biomechanics observed in humans.
These insights can improve realism in anthropomorphic robots and wearable structurally compliant robots.


%% file: acknowledgment.tex
\section{ACKNOWLEDGMENT}
We would like to thank NASA ARC UARC, the Advanced Studies Laboratories (ASL) and the Center for Information Technology for the Interest of the Society (CITRIS) for their support. 